\documentclass{article}

\PassOptionsToPackage{numbers, compress}{natbib}
\PassOptionsToPackage{table,xcdraw}{xcolor}
\usepackage[preprint]{neurips_2025}
 \usepackage{amsmath}
\usepackage{wrapfig}
\usepackage{algorithm}
\usepackage{algorithmic}
\usepackage[x11names,svgnames,dvipsnames]{xcolor} % 
\usepackage{xcolor}
\usepackage{multirow}
\usepackage[T1]{fontenc}    % use 8-bit T1 fonts
\usepackage{hyperref}       % hyperlinks
\usepackage{url}            % simple URL typesetting
\usepackage{booktabs}       % professional-quality tables
\usepackage{amsfonts}       % blackboard math symbols
\usepackage{nicefrac}       % compact symbols for 1/2, etc.
\usepackage{microtype}      % microtypography
\usepackage[most]{tcolorbox}
\usepackage{array} 
\usepackage{listings}
\usepackage{graphicx}  
\usepackage{subcaption} 
\usepackage{float}     
\usepackage{fontawesome5}
\usepackage{amssymb}

\usepackage{minted}

\usepackage{xcolor}

\setminted{
    autogobble,          
    numbersep=5pt,      
    frame=single,       
    framesep=2mm,        
    fontsize=\small,    
    breaklines,         
}

\usepackage{pifont}
\usepackage{array}
\usepackage{soul}
\usepackage{listings}
\usepackage{chngcntr}

\lstdefinestyle{prompt}{
  basicstyle=\ttfamily\footnotesize,
  breaklines=true,
  breakatwhitespace=false,
  columns=fullflexible,
  keepspaces=true,
  showstringspaces=false,
  frame=single,
  rulecolor=\color{black!30},
  backgroundcolor=\color{black!2},
  xleftmargin=0.6em,
  framexleftmargin=0.6em,
  framesep=0.4em
}

\AtBeginDocument{\AtBeginDocument{\counterwithout{lstlisting}{section}}}

\definecolor{lightblue}{RGB}{235,245,255} 
\definecolor{liautoblue}{RGB}{71,111,182} 
\definecolor{textred}{RGB}{128,0,0}
\usepackage{titlesec}
\titleformat{\section}
  {\large\bfseries\color{liautoblue}}{\thesection}{1em}{}
\titleformat{\subsection}
  {\bfseries\color{liautoblue}}{\thesubsection}{1em}{}
  \titleformat{\subsubsection}
  {\bfseries\color{liautoblue}}{\thesubsubsection}{1em}{}

\newtcolorbox{liautoabstract}{
    colback=lightblue,
    colframe=white,
    boxrule=0pt,
    arc=2mm,
    left=4mm,
    right=4mm,
    top=5mm,
    bottom=5mm,
    enhanced, 
    before upper={\setlength{\parindent}{0pt}} 
}
\newtcolorbox{stepTitle}[1]{
    enhanced,
    colback=gray!5,    
    colframe=black!50, 
    boxrule=-1pt,
    arc=0mm,            
    left=2mm, right=2mm, top=1mm, bottom=1mm,
    fontupper=\small,  
    title=#1
}

\tcbset{
    stepbox/.style={
        enhanced,
        colback=gray!20,         
        colframe=gray!50,        
        boxrule=0.5pt,
        title={#1},              
        fonttitle=\bfseries,    
        left=2mm, right=2mm, top=1mm, bottom=1mm
    }
}

\newtcolorbox[auto counter]{case}[2][]{ 
    enhanced,
    colback=gray!5,
    colframe=black!70,
    coltitle=white,
    fonttitle=\bfseries\sffamily,
    fontupper=\small,
    arc=1.5mm,
    boxrule=0.5pt,
    title=Case study \thetcbcounter: #2, 
    left=1mm, right=1mm, top=2mm, bottom=2mm,
  
    label type=case, 
    #1              
}

\newtcolorbox{toolbox}[1]{
    enhanced,                 
    colback=gray!5,          
    colframe=black!70,        
    coltitle=white,           
    fonttitle=\bfseries\sffamily,
    fontupper=\small,
    arc=1.5mm,              
    boxrule=0.5pt,            
    title=#1,                 
    left=1mm, right=1mm, top=2mm, bottom=2mm
}

\usepackage{fancyhdr}
\usepackage{graphicx} 

\hypersetup{
    colorlinks=true,
    urlcolor=liautoblue,
    citecolor=liautoblue,   
}
\newcommand{\eg}{\textit{e.g.}}
\newcommand{\ie}{\textit{i.e.}}

\usepackage{enumitem}

\title{StreamingClaw Technical Report}

\author{%
  MindGPT-ov Team \\
   Li Auto Inc.   \\
}

\begin{document}

\maketitle

\begin{liautoabstract} 

Emerging applications such as embodied intelligence, AI hardware, autonomous driving, and intelligent cockpits rely on a real-time perception-decision-action closed loop, posing stringent challenges for streaming video understanding. However, current agents mostly suffer from fragmented capabilities, such as supporting only offline video understanding, lacking long-term multimodal memory mechanisms, or struggling to achieve real-time reasoning and proactive interaction under streaming input. These shortcomings have become a key bottleneck for preventing agents from sustaining perception, making real-time decisions, and executing closed-loop actions in complex real-world environments, constraining their deployment and potential in dynamic, open physical worlds. 
To alleviate these issues, we propose \textbf{StreamingClaw}, a unified agent framework for streaming video understanding and embodied intelligence. Beyond maintaining full compatibility with the OpenClaw framework, it natively supports real-time, multimodal streaming interactions. StreamingClaw integrates five core capabilities: (1) It supports real-time streaming reasoning. (2) It supports reasoning about future events and proactive interaction under the online evolution of interaction objectives. (3) It supports multimodal long-term memory storage, hierarchical memory evolution, efficient memory retrieval, and memory sharing across multiple agents. (4) It supports a closed loop of perception-decision-action. In addition to conventional tools and skills, it also provides streaming tools and action-centric skills tailored for real-world physical environments. (5) It is compatible with the OpenClaw framework, allowing it to leverage the resources and support of the open-source community. With these designs, StreamingClaw integrates online real-time reasoning, multimodal long-term memory, and proactive interaction within a unified framework. Moreover, by using scalable tools and skills to translate decisions into executable actions, it enables direct control of the physical world, supporting practical deployment of embodied interaction. The project page is at \href{https://jackyu6.github.io/StreamingClaw-Page/}{StreamingClaw}.

\end{liautoabstract}

\section{Introduction}
Embodied intelligence as physical actors (robots \cite{duan2022survey, li2025clivis, long2025survey, xiong2024aic}, autonomous driving \cite{gao2026foundation, zhang2024ad}, and embodied agents \cite{fung2025embodied, wu2025generative}) rely on video streams as one of the primary perceptual inputs. Therefore, it should support low-latency streaming video understanding with continuous spatiotemporal perception. Otherwise, it may suffer from action delays or make hasty decisions without leveraging long-horizon information, directly leading to task failure. Current real-time streaming video understanding faces the following challenges:

\textbf{(1) Streaming perception}. Real-world environments are not non-stationary and continuously evolving (with people, objects, and scenes moving dynamically). They cannot be treated as offline videos for pre-processing. Instead, the embodied intelligence system should rely on streaming, incremental methods to perceive the constantly updated state of the environment, which is an essential prerequisite for deploying embodied intelligence in real-world scenarios.

\textbf{(2) Long-term memory}. Streaming input is inherently a continuous spatiotemporal representation of the physical environment, carrying key information about its dynamic evolution. Embodied intelligence should depend on long-term memory to build a comprehensive, dynamic, and effective understanding. If an agent relies only on the limited frames or short video clips for local perception, its interaction capability and task-execution reliability will degrade substantially \cite{wang2025mirixmultiagentmemoryllmbased}.

\textbf{(3) Proactive interaction}. A core requirement of real-time streaming video understanding for embodied intelligence is to directly translate visual semantic information into executable action commands, enabling seamless coupling between perceptual input and action execution. This requires moving beyond the limitations of passive perception and leveraging active perception to acquire environmental information accurately and efficiently, thereby providing effective support for decision-making and action, which is an important prerequisite for autonomous execution of complex tasks.

To address the above challenges, several recent works have adopted certain approaches:
For streaming perception, existing works leverage visual compression \cite{tao2025dycoke, chen2025streamingtom} or visual token selection \cite{hu2025m, tang2025adaptive} to reduce redundancy across sequential frames. However, critical fine-grained information is often lost during compression and selection, making it difficult to reliably memorize and retrieve historical content. Therefore, a more reliable agent framework is needed, which maintains low-latency responsiveness while providing more robust memory support.
For long-term memory, existing works rely on the model's native context awareness \cite{xiong2025streaming, yang2025streamagent}. They remember historical actions and events in current context, giving the model a certain degree of long-horizon perception. However, the memory constructed in this way is highly limited, as it typically contains only textual information or historical KV cache and cannot support human-like, vision-scene-based recall. As interaction time grows, this approach accumulates substantial redundancy and struggles to focus on the important information. 
For proactive interaction, existing works often use salient changes in the visual stream as triggers to activate the model \cite{yao2025timechat}, other works introduce lightweight modules to decide whether the agent should respond proactively, triggering responses by learning such signals \cite{wang2025streambridge, qian2025dispider}. However, these modules typically rely on heuristic rules and have limited capability for complex context understanding and long-horizon dependency modeling.

The above approaches alleviate several key issues in streaming video understanding to some extent, but their limitations remain clear. On the one hand, they are still insufficient to systematically cover and address the three core challenges outlined above. On the other hand, most existing methods remain at the level of the model's perception and understanding, lacking the ability to further translate understanding into executable policies that can drive real actions and alter the physical world.
To this end, we propose the StreamingClaw framework, which represents video streams as continuous spatiotemporal data and addresses the above three core challenges through an autonomous multi-agent scheduling mechanism. It also flexibly integrates a rich suite of tools and skill libraries, enabling instruction-driven embodied intelligence in real-world scenarios. The recently proposed agent framework OpenClaw \cite{OpenClaw} likewise provides strong human–computer interaction and practical problem-solving capabilities. However, it is primarily designed for static, text-based interaction. In contrast, StreamingClaw is tailored to real-time streaming and dynamically changing embodied interaction scenarios. Moreover, StreamingClaw is compatible with OpenClaw’s capabilities, making it applicable to a broader range of settings.
The core functionalities of StreamingClaw are as follows:

\textbf{(1) Main-sub-agent collaborative framework.} 
StreamingClaw standardizes and structures multi-end inputs, transforming them into a unified representation that can drive decision-making at the agent layer. It adopts a main–sub agent collaborative architecture, where the main agent \textbf{StreamingReasoning} is designed to be compatible with outstanding open-source multimodal models \cite{bai2025qwen25vltechnicalreport, qwen3-vl, qwen3.5, mindgpt4ov, chen2025mindwatcher} and serves as StreamingClaw’s core decision agent. It performs incremental understanding and streaming reasoning at the frame level, enabling real-time watch-and-respond interactions. Meanwhile, it can autonomously plan tasks and delegate them to appropriate sub-agents to handle specific subtasks.

\textbf{(2) Scalable sub-agents.} StreamingClaw employs lightweight sub-agents compatible with streaming video understanding to meet scenario-specific requirements and support the main agent's decision-making. First, sub-agent \textbf{StreamingMemory} addresses challenges where context continuously accumulates and past information cannot be replayed under streaming video understanding. It provides traceable multimodal memories and supports hierarchical evolution from short-term memory to long-term memory. Through a flexible dynamic evolution mechanism and efficient retrieval capabilities, it enables multimodal memory modeling across different agents. Second, sub-agent \textbf{StreamingProactivity} makes proactive interaction decisions. It targets complex real-world applications and supports completing generalizable and proactive goals that can adapt dynamically to changing conditions.

\textbf{(3) Perception–decision–action closed loop.} While supporting a wide range of community tools and skill libraries, StreamingClaw targets streaming video understanding scenarios and builds interactive tools and skills that can solve real-world physical problems. It supports the final link from perception to decision-making and then to physical action, ultimately enabling an embodied agent to achieve a closed-loop interaction cycle in the real physical world.

\begin{figure}
    \centering
    \includegraphics[width=\textwidth]{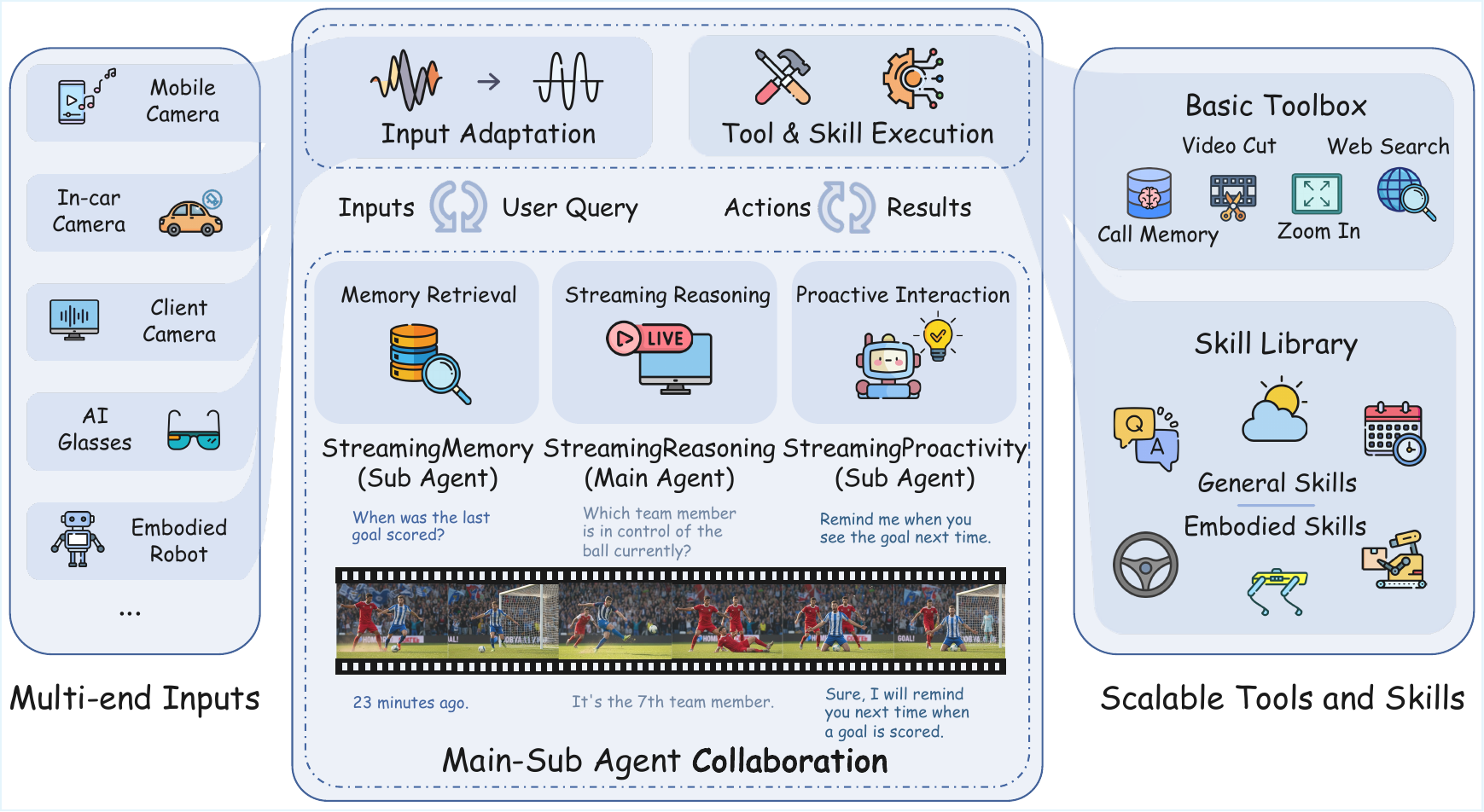}
    \caption{Pipeline of StreamingClaw: Multi-end inputs and the user query are fed into the main–sub agents for perception and decision-making. The instructions produced by the main–sub agents then guide the execution of tools and skills, whose results are fed back to the main–sub agents, forming a closed-loop pipeline of perception–decision–action.}
    \label{fig:framework_cropped}
\end{figure}

\section{Framework of StreamingClaw}
This chapter presents an overview of StreamingClaw’s architecture and execution pipeline. It first explains its input access mechanism for multi-end data sources and the unified representation method and then describes how adaptive support for different terminals is achieved on this basis (see Sec.\ref{sec: multi-end-input}). Next, it introduces the end-to-end execution pipeline from input to output (see Sec.\ref{sec:pipeline}), including how the modules of StreamingClaw collaborate and how they are connected throughout the streaming video inference.

\subsection{Multi-end Input Adaptation}
\label{sec: multi-end-input}
StreamingClaw can process streaming inputs from multiple types of devices, including handheld devices, vehicles, smart glasses, and embodied robots. Such streaming inputs provide a continuous spatiotemporal representation of the physical environment. Compared with unimodal data, they are more complex in form and more resource-intensive. To accommodate multimodal streaming inputs, we make the following adaptations:

\begin{itemize} [leftmargin=*, itemindent=0.0em]
\item \textbf{Input standardization.} StreamingClaw applies a unified standardization pipeline to streaming inputs from different endpoints: aligning them by timestamps and obtaining absolute time via anchors, which facilitates subsequent interactions between the main agent and sub-agents. To improve StreamingClaw’s performance across different devices, we optimize it based on the native data quality on each endpoint, provide a configurable parameter table, and adjust these parameters during runtime according to feedback from the results.

\item \textbf{Shared streaming cache.} To reduce StreamingClaw’s computational resource usage, we adopt a shared streaming cache queue and set its maximum length according to scenario requirements. Taking the video-frame cache as an example, the main agent and sub-agents share the same cache resources. The cache queue supports different application scenarios along two dimensions: time window and frame density.
For the time-window dimension, it provides relatively long chunks to meet low-frequency inference demands, such as long-term perception tasks in a silent state that do not require rapid feedback. It also provides short chunks for tasks requiring fast responses, such as high-frequency memory updates and real-time proactive reminders.
For the frame-density dimension, it can store fast frames and slow frames, which respectively serve tasks that require long-term perception and instantaneous perception.

\item \textbf{Dynamic prompt construction.} When user queries are received from multiple endpoints, they are fed into StreamingClaw via dynamic prompt construction to coordinate the collaboration and execution of different agents. In addition to maintaining compatibility with OpenClaw’s prompt-construction logic \cite{OpenClaw}, we introduce targeted modifications to better support multimodal streaming interactions. For example, StreamingClaw continuously maintains absolute timestamps in the main agent’s prompt, providing an absolute temporal scale for invoking various sub-agents and tools and for temporal coordination. For proactive responses, upon receiving a user query, StreamingClaw decomposes and generalizes the user’s proactive intent. For streaming memory, StreamingClaw supports the evolutionary collaboration between short-term and long-term memories.
\end{itemize}

\subsection{Pipeline of StreamingClaw}
\label{sec:pipeline}
This section introduces the multi-agent collaborative execution pipeline of StreamingClaw for streaming inputs (see Fig.\ref{fig:framework_cropped}).
StreamingClaw first receives streaming inputs from multiple endpoints and performs standardized and structured processing over these inputs. By parsing and transcribing the signals collected at each endpoint, it forms a standardized multimodal streaming data representation, providing a stable and unified input for subsequent real-time inference and interaction. The processed inputs are then fed into the main agent for streaming reasoning, which generates corresponding output signals. Based on the output, the system determines whether tools or skills need to be invoked. If not, the result is returned directly to the user.

In this pipeline, StreamingReasoning is responsible for real-time streaming perception and planning, while making overall interaction decisions by incorporating feedback from sub-agents (see Sec.~\ref{sec:streaming_mind}). The sub-agent StreamingMemory is responsible for building and managing multimodal memory to support the main agent's decisions (see Sec.~\ref{sec:hierarchical_memory}). The StreamingProactivity agent is responsible for proactive interaction decision-making (see Sec.~\ref{sec:proactive_mind}). 
Finally, when the agents output action instructions, the toolbox and skill library execute the corresponding tools and skills (see Sec.~\ref{sec:tool_and_skill}). Tools are mainly used for single-step actions with clear goals and simple execution, covering basic text-image, video, and memory-related tools. Skills are designed to handle complex action sequences, including general daily-life and entertainment skills as well as embodied-interaction skills.

\begin{figure}
    \centering
    \includegraphics[width=\textwidth]{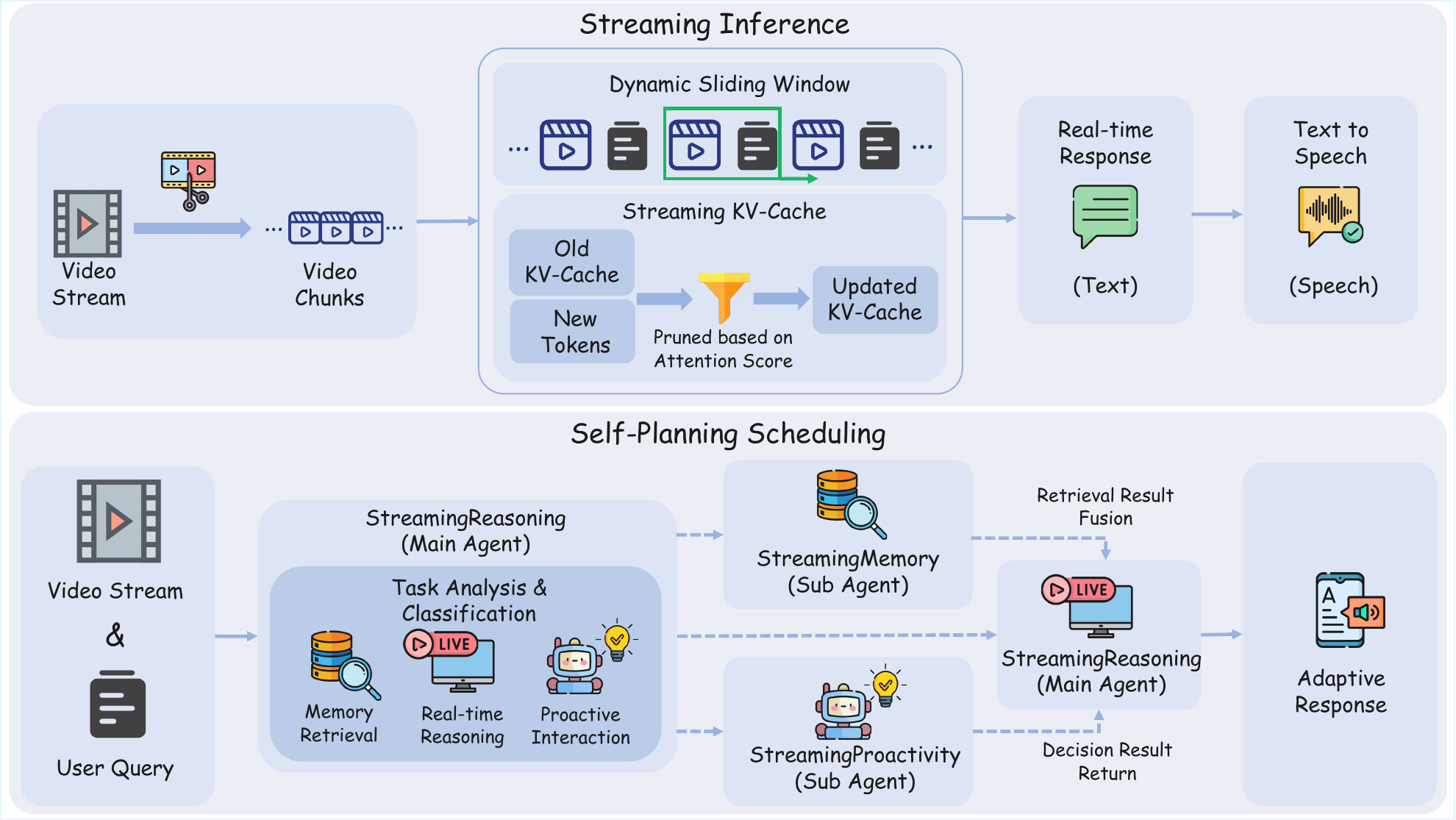}
    \caption{Flowchart of SteamingMind’s streaming inference and self-planning scheduling.}
    \label{fig:StreamingReasoning_Agent_cropped}
\end{figure}

\section{StreamingReasoning Agent}
\label{sec:streaming_mind}
StreamingReasoning targets streaming video understanding scenarios with continuous input and output. Its main goal is to achieve real-time perception, understanding, and reasoning under low-latency constraints, responding to user queries and generating results in real time. Inspired by works on streaming reasoning \cite{tao2025dycoke, xu2025streamingvlm}, StreamingReasoning maintains a dynamically updated KV-Cache \cite{kvsurvey} pool during inference and adopts a dynamic sliding window mechanism to retain only the visual and textual context within the most recent time window, thereby controlling context length and GPU memory overhead for long-duration video streams.

At the computational level, to avoid the high cost of recomputing over the entire history at every step, StreamingReasoning reuses cached KV tokens in each incremental inference step and computes only the incremental tokens introduced by newly arrived chunks, achieving stable throughput and low latency. Meanwhile, StreamingReasoning prunes tokens in the KV-cache: based on attention-based contribution scores, it removes cached tokens with low scores, further reducing attention computation and overall inference overhead.

In addition to streaming inference, StreamingReasoning also serves as the main agent responsible for multi-agent scheduling. StreamingReasoning parses user instructions and determines the task type (\eg, whether historical memory is needed, whether the task involves proactive interaction decisions, or whether tool or skill invocation is required), and then autonomously plans the execution workflow accordingly.

By combining techniques described above, StreamingReasoning delivers a streaming reasoning and interaction experience close to watch-and-answer under continuous streaming videos while keeping latency and computational cost controllable over long-running sessions.

\subsection{Streaming Inference}
The core idea of streaming inference in StreamingReasoning is to transform an offline video agent into an online version that supports streaming inputs and outputs, enabling it to continuously receive streaming videos, perform streaming inference, and generate responses with low latency. 
The top part of Fig.~\ref{fig:StreamingReasoning_Agent_cropped} illustrates the overall flow of the streaming inference. Specifically, StreamingReasoning segments the incoming streaming videos into fine-grained temporal chunks along the time axis: within each time window, a certain number of frames are sampled as the processing unit. Whenever a new chunk arrives, StreamingReasoning performs one round of encoding and inference.
To control context length and computational overhead under long-duration inputs, StreamingReasoning introduces a dynamic sliding window mechanism: the window slides forward over time, retaining only the visual and textual context within the most recent time range. Information outside the window is discarded according to a pre-defined policy or offloaded to the memory agent, thereby preventing unbounded context growth.

In terms of inference acceleration, StreamingReasoning further introduces a streaming KV-Cache. By reusing the KV-Cache computed from previous steps, each decoding step only needs to compute the incremental part for newly arrived tokens, rather than repeatedly recomputing all historical tokens. This enables low latency and stable throughput in long-horizon streaming video understanding.
Meanwhile, to further reduce GPU memory usage and attention computation, the KV-Cache is pruned. During LLM decoding, a set of Transformer layers  \cite{vaswani2017attention} is selected to compute attention scores between cached tokens and newly input tokens. The scores corresponding to the visual modality are then identified and used for selective pruning. At each decoding step, only high-contribution visual tokens are retained in the KV-Cache, while the cache state is updated accordingly. The overall procedure mainly consists of the following three steps:

\begin{itemize} [leftmargin=*, itemindent=0.0em]
\item \textbf{Step1:} In the first decoding iteration, the initial batch of visual tokens is first written into the KV-Cache, and the cross-attention weights between the predicted tokens and the visual tokens at the \(L\)-th layer are calculated. Subsequently, sorting is performed based on the attention scores, and the visual tokens with scores ranking in the top \(p\%\)  are selected as high-importance tokens, with their corresponding Key-Value pairs retained in the KV-Cache.

\item \textbf{Step2:} In subsequent decoding iterations, when the variation amplitude of the visual scene is small, new visual tokens may not be written into the KV-Cache to further improve the effective utilization rate of tokens during the decoding phase. Specifically, the cosine similarity between the input visual tokens and cached tokens is calculated. If the similarity is higher than a preset threshold, their information is judged to be redundant and the cache update is skipped. The remaining visual tokens are allowed to be written into the cache.

\item \textbf{Step3:} Once new visual tokens need to be written into the KV-Cache, the pruning and updating procedure in Step1 is repeated. That is, the visual tokens in the cache are sorted and pruned according to the cross-attention scores: tokens with scores in the top \(p\%\) are regarded as high-importance tokens, which continue to participate in subsequent attention computations and remain in the KV-Cache; the remaining tokens are removed from the KV-Cache to control the cache size and reduce computational and memory-access overhead.

\end{itemize}
The aforementioned pruning and caching mechanism runs iteratively at each decoding step. While preserving key information and contextual consistency, low-importance visual tokens are dynamically filtered and discarded to avoid involvement in subsequent Transformer decoding, which notably reduces attention computation complexity and KV-Cache read–write overhead, thus boosting overall inference efficiency.

By combining the dynamic sliding window and streaming KV cache, StreamingReasoning transforms the computation pattern from repeated full-history recomputation that grows over time to linear incremental updates over newly arriving inputs. This enables the agent to maintain a stable, real-time understanding and reasoning under long, continuous streaming videos, achieving an effect close to watching and answering simultaneously.
Furthermore, by integrating components ASR (automatic speech recognition) \cite{jain2025sage,2024automatic} and TTS (text-to-speech synthesis) \cite{tss}, StreamingReasoning can extend text-level outputs to real-time spoken interaction, enabling a watching-and-speaking or listening-and-answering experience. 

\subsection{Self-Planning Scheduling}

As the main agent, StreamingReasoning determines the task type of the user query based on a dynamically constructed system prompt, then autonomously plans and schedules sub-agents. We illustrate its self-scheduling process in the bottom part of Fig.~\ref{fig:StreamingReasoning_Agent_cropped}. The detailed overall workflow is as follows:

\begin{itemize}[leftmargin=*, itemindent=0.0em]
\item \textbf{Task parsing and categorization.} StreamingReasoning first performs semantic parsing of the user query, determines whether it falls into memory-augmented retrieval, real-time understanding and reasoning, or proactive interaction decision-making. Then, StreamingReasoning assesses whether collaboration with sub-agents is required.

\item \textbf{Memory retrieval path (if needed).} If the task is judged to depend on historical information or personalized context (\eg, user preferences, prior dialogue conclusions, historical events), StreamingReasoning invokes the `call\_memory` tool in the toolbox to dispatch the sub-agent StreamingMemory for hierarchical retrieval. It then fuses the retrieved results with newly observed online streaming video information to form a unified context, followed by reasoning, decision generation, and output.

\item \textbf{Proactive interaction decision path (if needed).} If the query is further classified as a proactive interaction decision task (\eg, proactive reminders, proactive summaries, or anomaly alerts), StreamingReasoning assigns it directly to the sub-agent StreamingProactivity. StreamingProactivity combines the real-time state and interaction strategy to decide when to intervene, what to output, and what form to use (prompt, follow-up question, summary, warning, etc.), then returns the result to the main agent, StreamingReasoning, for unified orchestration and delivery to the user.

\item \textbf{No-memory and no-proactive-interaction path.} If the task neither depends on historical memory nor requires proactive interaction and decision-making, StreamingReasoning performs understanding and reasoning directly based on the streaming multimodal information within the current window, reducing additional overhead and ensuring low response latency.

\end{itemize}

Through this self-planning scheduling mechanism, StreamingReasoning achieves adaptive scheduling: invoke memory when needed, proactively interact when needed, and otherwise perform direct streaming reasoning, which improves real-time reasoning and user interaction experience.

\begin{figure}
    \centering
    \includegraphics[width=\textwidth]{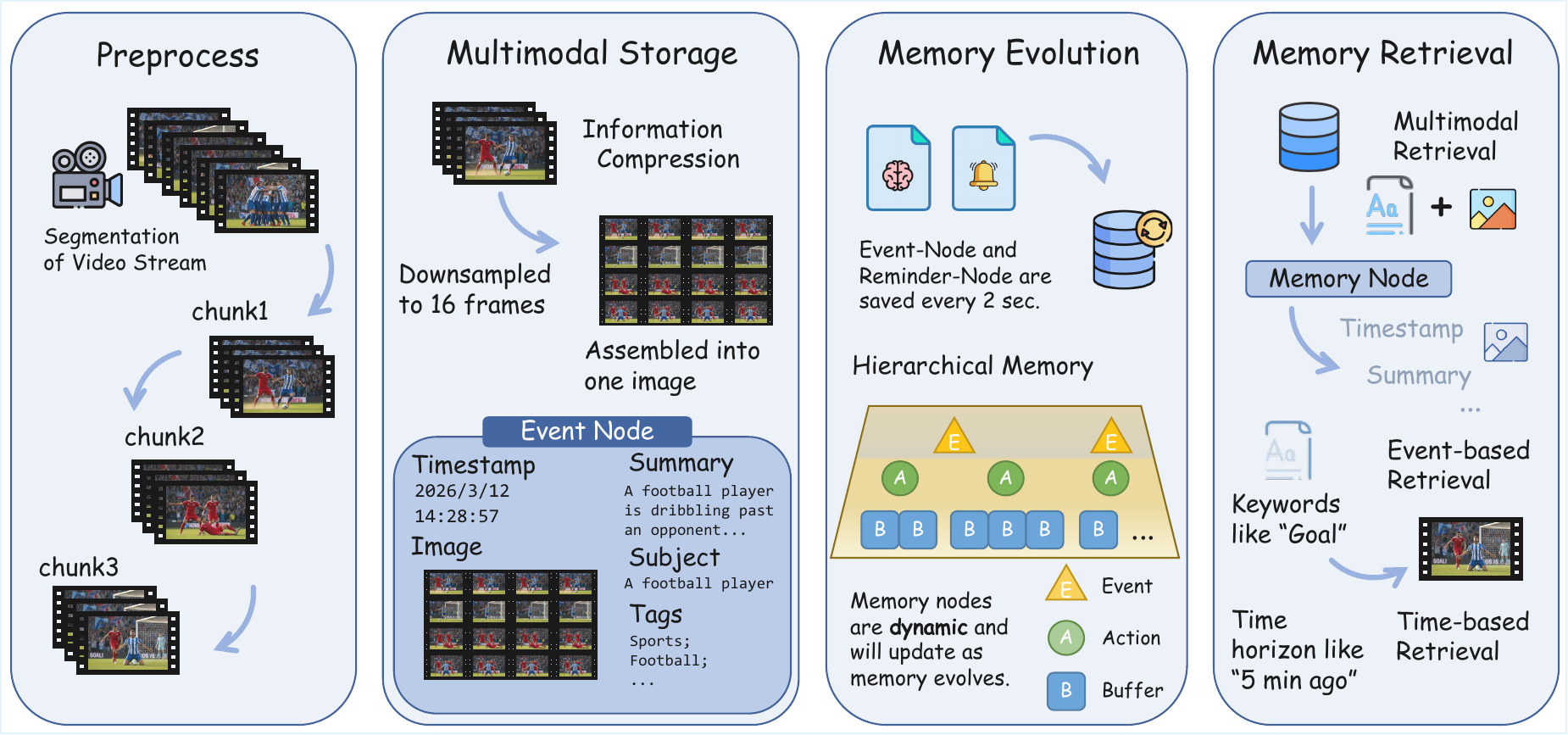}
    \caption{Flowchart of StreamingMemory’s memory storage, evolution, and retrieval algorithm process.}
    \label{fig:memory_agent_cropped}
\end{figure}

\section{StreamingMemory Agent}
\label{sec:hierarchical_memory}
  
Real-time streaming videos are continuous, highly redundant, and strongly temporal data in which multiple modalities coexist. Streaming video agents typically use a sliding window to perceive local information over a short time span. This approach can cause the agent to lose important contextual information, leading to knowledge fragmentation when global reasoning is required. A common practice is to use a memory system to directly insert stored memories into the context as supplementary information for model reasoning \cite{chhikara2025mem0buildingproductionreadyai, wang2025mirixmultiagentmemoryllmbased}. However, this approach has the following drawbacks:

\begin{itemize} [leftmargin=*, itemindent=0.0em]
\item \textbf{Information loss.} Memory is stored only in textual form. In multimodal interaction settings, this can lead to semantic misalignment and loss \cite{liu2026simplememefficientlifelongmemory, long2025seeinglisteningrememberingreasoning}.

\item \textbf{Inefficiency.} Long-term interactions under streaming video understanding generate a large amount of memory. Directly injecting it into the context degrades inference efficiency and consumes excessive tokens. 
If instead it retrieves only important information to inject, it faces challenges in retrieval effectiveness and efficiency.

\item \textbf{Rigid memory.} Each memory entry is fragmented and isolated, lacking logical connections between different pieces of information and no distinction in importance. When all memories are treated with equal weight, their relative importance cannot be reflected in the model's reasoning outputs, ultimately making memory usage rigid and inflexible \cite{long2025seeinglisteningrememberingreasoning}.
\end{itemize}

Therefore, how to effectively acquire, store, organize, and utilize memory has become a major challenge in streaming video understanding. To address the above issues, we introduce a streaming memory agent, StreamingMemory. We illustrate its algorithmic workflow in Fig.~\ref{fig:memory_agent_cropped}, its core mission is to store, evolve, and retrieve multimodal information from streaming videos in real time, providing long-term contextual support for complex reasoning tasks such as temporal question answering or logical relation inference. Unlike traditional memory systems that only offer simple storage and retrieval as discussed above, StreamingMemory introduces multimodal memory representations and an active evolution mechanism, enabling hierarchical, flexible, efficient, and semantically informed memory management. Overall, StreamingMemory has the following characteristics:

\begin{itemize} [leftmargin=*, itemindent=0.0em]
\item \textbf{Multimodal memory storage.} Semantic information is stored as embedding vectors, supporting both short-term and long-term memory. Short-term memory is more informative, capturing events, actions, trajectories, and states from the most recent seconds or minutes, while long-term memory is more abstract and condensed.
\item \textbf{Flexible evolution.} Memories can be added, updated, or deleted, ensuring important information receives sufficient emphasis while reducing interference from redundant content.
\item \textbf{Efficient retrieval.} Reducing computational overhead and mitigating error accumulation caused by multi-round retrieval.
\item \textbf{Unified memory across agents.} 
Memory storage and retrieval are unified across agents, while memory management is differentiated across agents.
\end{itemize}

\subsection{Multimodal Memory Storage}
Most traditional memory systems forcibly convert visual information into text for storage \cite{chhikara2025mem0buildingproductionreadyai}, which leads to severe semantic misalignment and information loss. In StreamingMemory, memory is no longer limited to textual summaries and descriptions, it instead becomes a vision-centric multimodal memory. Within continuously arriving multimodal streaming inputs, historical information is written into the memory repository in an incrementally updatable, retrievable, and evolvable form, providing a stable contextual foundation for long-term reasoning. To enable this, we define a memory node as the basic organizational unit of memory. At time 
$t$, the agent receives the input video segment and constructs the following memory node:
${n_t = (z, s, c, \tau)_{t}}$, 
$z$ is the compressed video segment, $s$ represents the textual summary of the segment, 
$c$ represents the detailed description of this segment, and $\tau$ denotes the timestamp at which the segment ends.

\subsection{Hierarchical Memory Evolution}
To mitigate the bottlenecks of long-term memory in event consistency and retrievability, we introduce a Hierarchical Memory Evolution (HME) mechanism. When continuously receiving streaming videos, HME enables the memory agent to perform memory summary, redundancy removal, and structured compression in a humanlike manner. Based on the timeliness and level of summary of the information, memories evolve into short-term memories and long-term memories. The core idea is to construct a multi-level summary hierarchy in the memory representation space: segments $\to$ atomic actions $\to$ events, so that short-term, fine-grained memories are progressively evolved into long-term, stable, structured memories, thereby improving the semantic accessibility and robustness of memory retrieval.

Short-term memory mainly consists of compressed video frames and simply summarized atomic actions, which are fine-grained and clear. 
However, due to the lack of contextual logic, a single short-term memory often cannot capture the intent behind an action or its temporal dependencies with preceding and subsequent events. As the hierarchical evolution mechanism progresses, fragmented short-term memories are aggregated into event-level long-term memory. After this evolution, the long-term memory is no longer a collection of discrete frames but highly compressed and structured. Long-term memory is relatively abstract and coarse in representation, yet it can extract macro-level information with strong semantics and context cues.

Specifically, StreamingMemory first performs segment $\to$ atomic-action induction. For each newly input video segment, StreamingMemory retrieves a set of neighboring memory nodes from the current memory repository and defines a compatibility score for merging into an atomic-action node, jointly accounting for semantic similarity and temporal continuity. An online update is then conducted via a thresholding policy: if the score exceeds a predefined threshold, the segment is merged into the corresponding atomic action. Otherwise, a new atomic-action node is created. This step consolidates fragmented, repetitive fine-grained segments into semantically coherent atomic actions, effectively reducing redundancy and noise.

Subsequently, HME performs atomic-action $\to$ event aggregation. Given a temporally contiguous sequence of induced atomic actions, StreamingMemory abstracts them into higher-level event nodes. Event formation is governed by a consistency constraint over the same objects and scenes, which can be formalized as an aggregation criterion based on scene similarity. Concretely, a scene-similarity score is computed, and if the merging condition is satisfied, the existing event is updated; otherwise, a new event is created.

Through the above online induction and merging mechanisms, HME progressively abstracts streaming inputs from granular fragments into atomic actions and events, achieving the following:

\begin{itemize} [leftmargin=*, itemindent=0.0em]
\item \textbf{Temporal queryability.} Temporal queryability: Within each event, the atomic-action chain explicitly preserves temporal order, improving the robustness of temporal understanding and retrieval over long streaming videos.

\item \textbf{Redundancy compression.} Repetitive segments are merged into atomic actions, which are further aggregated into events, mitigating memory fragmentation.

\item \textbf{Structured long-term storage.} Events serve as stable memory chunks, enhancing the scalability of memory storage.
\end{itemize}

\subsection{Efficient Memory Retrieval}
To enable efficient memory retrieval, we propose a retrieval method that is compatible with command-driven control, high-concurrency, and self-directed temporal traversal. Its core goal is to reduce the computational overhead of unproductive traversal without sacrificing retrieval quality, while also mitigating the error accumulation that arises from multi-round serial reasoning.
Specifically, it includes the following retrieval strategies:

\begin{itemize} [leftmargin=*, itemindent=0.0em]
\item {\textbf{Command-driven retrieval.}} StreamingMemory no longer acts merely as a passive candidate matcher. It instead serves as the strategy controller of the retrieval process. Based on the question type and difficulty (\eg, single-fact recall vs. temporal inference), it adaptively determines the search depth and stopping criteria. 
In each retrieval round, it decides whether to continue traversing or to stop early and output, enabling early stopping once relevant memories are found and avoiding unnecessary deep retrieval and full-database scanning.

\item {\textbf{High-concurrency retrieval.}} Candidate-memory matching, re-ranking, and evidence extraction are processed in parallel. On the one hand, parallelization improves system throughput and reduces end-to-end latency. On the other hand, it avoids the chain propagation of bias in serial multi-round generative interactions, where outputs from earlier rounds influence later judgments, thereby reducing the accumulation of errors.

\item {\textbf{Self-directed temporal traversal.}} To further improve early-hit capability, the StreamingMemory can autonomously choose the traversal order based on question requirements and memory structure, including (1) forward traversal to preserve temporal and causal consistency; (2) reverse traversal to prioritize the most recent information; and (3) salience-first traversal, where each memory is assigned a salience score at write time and salient segments are accessed first during retrieval. This strategy allows candidate-set construction and resource allocation in concurrent retrieval to focus on high-value memories earlier, improving effective memory recall and response speed under a limited compute budget.
\end{itemize}

In summary, StreamingMemory adopts a command-driven, high-concurrency, self-directed temporal traversal retrieval strategy. StreamingMemory dynamically sets stopping criteria, obtains candidates through concurrent execution, and uses temporally salience-driven traversal to achieve higher early-hit rates with less wasted computation, thereby striking a better balance between efficiency and robustness.

\subsection{Cross-Agent Unified Memory}

In a multi-agent collaborative framework, a memory system designed for a single agent can no longer meet the multidimensional requirements of complex tasks. StreamingMemory introduces a cross-agent unified approach to memory storage and retrieval, aiming to address the interface heterogeneity that arises during multi-agent collaboration. This allows memory storage and retrieval to be standardized at the cross-agent level, while preserving differentiation in memory management and output information.

To provide a unified memory interface in a cross-agent collaborative framework, we first standardize the methods for storage and retrieval. Although different agents take on different tasks, their interaction interfaces follow strict structural specifications, and all memory nodes adhere to a unified structured definition. StreamingMemory provides standardized memory invocation methods, enabling any agent to write acquired multimodal information into the memory repository in the same way and can also obtain retrieval results with the same structure through a unified, normalized retrieval interface.

Meanwhile, StreamingMemory introduces differentiated memory management mechanisms for different agents to enable finer-grained control. For example, the memory node set required by StreamingReasoning emphasizes the global semantic evolution of video inputs and complex logical induction, supporting cross-temporal question answering and responding to user queries. In contrast, the memory node set required by StreamingProactivity focuses on key action triggers and monitoring state changes.

\begin{figure}
    \centering
    \includegraphics[width=\textwidth]{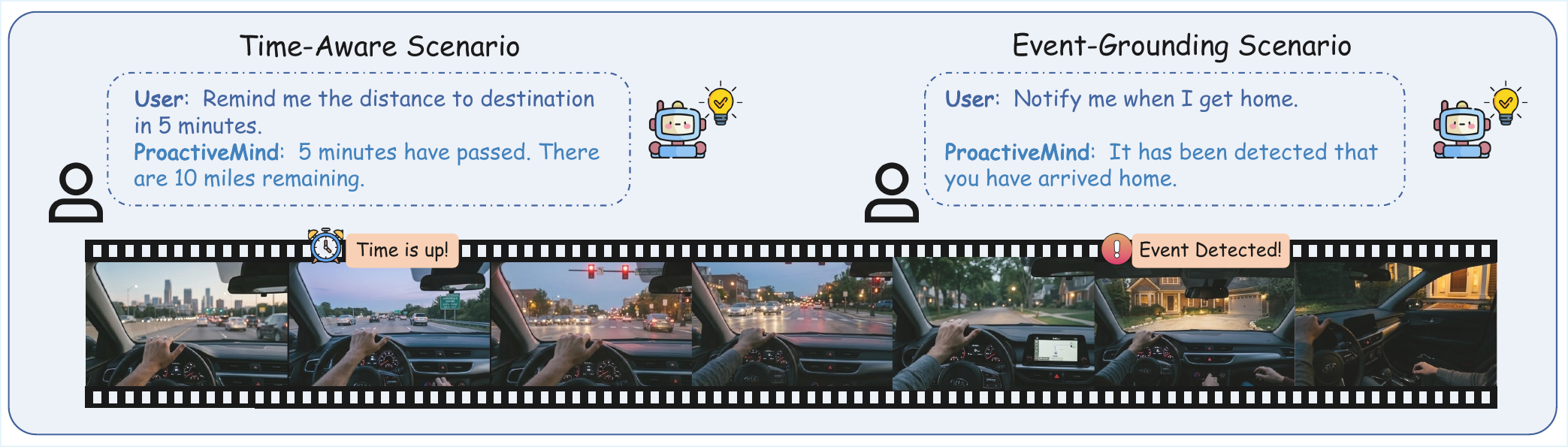}
    \caption{Interaction scenarios of the StreamingProactivity agent: time-aware and event-grounding.}
    \label{fig:StreamingProactivity_Scenario_cropped}
\end{figure}

\section{StreamingProactivity Agent}
\label{sec:proactive_mind}

StreamingProactivity is designed for future-event prediction, reasoning, and proactive interaction. Its proactive objectives can be specified upfront by the user or continuously evolve during streaming interaction. Specifically, when a user query is deemed a proactive interaction query, the main agent dispatches an instruction to StreamingProactivity, converting the focus into a continuously active online monitoring task (\eg, tracking the subsequent behavior of a specific object, determining whether a certain event occurs, monitoring whether risk conditions are triggered, etc.). Once StreamingProactivity captures evidence that satisfies the specified conditions, such as an event occurrence, a state reaching a threshold, or a significant change in a behavioral trend, it proactively generates a notification, alert, or explanatory response under the main agent’s unified scheduling. This eliminates the need for repeated user follow-ups and forms a closed loop of perceive $\to$ reason $\to$ trigger $\to$ feedback.
As shown in Fig.~\ref{fig:StreamingProactivity_Scenario_cropped}, StreamingProactivity supports two typical proactive interaction scenarios:

\begin{itemize} [leftmargin=*, itemindent=0.0em]

\item \textbf{Time-aware interaction.} This scenario emphasizes continuous state evolution over time. The goal is to characterize dynamics and long-term dependencies along the video timeline, which requires memorizing and reasoning over global temporal information. For example, in a driving scenario, given the user query: ``Remind me of the distance to the destination in 5 minutes,'' the agent must continuously track trip progress, then proactively trigger a reminder at an appropriate time with a reliable estimate.

\item \textbf{Event-grounding interaction.} This scenario focuses on key events of user interest. The goal is to accurately ground the time point (or interval) of an event and its context within a temporal stream. This requires high sensitivity to local changes and fine-grained cues while leveraging surrounding context for verification and attribution. Typical applications include anomaly detection (\eg, abnormal-behavior alerts in surveillance) and event annotation (\eg, automatic highlight clipping in sports).
\end{itemize}

To implement StreamingProactivity, we propose two distinct paradigms: a training-free and a training-based approach. The former requires no additional training and is seamlessly compatible with existing open-source models, thereby facilitating rapid deployment and practical implementation. However, its generalization capabilities are relatively constrained when confronted with complex scenarios and cross-domain tasks. Conversely, the latter achieves stronger adaptability and generalization through dedicated training, but at the expense of additional data requirements and computational overhead. 

\subsection{Training-free Adaption}
\begin{figure}
    \centering
    \includegraphics[width=\textwidth]{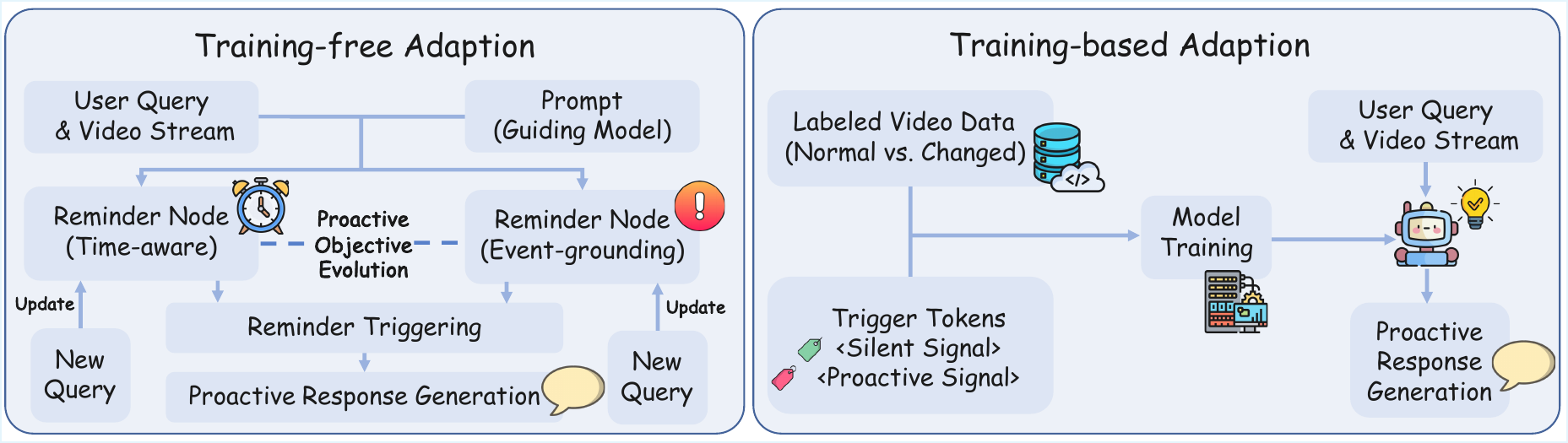}
    \caption{Implementation of the StreamingProactivity agent: training-free adaptation and training-based adaptation.}
    \label{fig:StreamingProactivity_Agent_Schemes_cropped}
\end{figure}

The training-free adaptation approach does not introduce additional model training yet can generalize to proactive interaction objectives and support real-time objective evolution in streaming interactions. As shown in the left part of Fig.~\ref{fig:StreamingProactivity_Agent_Schemes_cropped}, the core workflow consists of three stages: reminder node generation, proactive response matching, and proactive objective evolution.

\textbf{(1) Reminder node generation.} When StreamingProactivity receives a preset proactive objective or a proactive interaction request initiated from the user during video progress, it structurally encodes these active trigger conditions into sustainable, monitorable reminder nodes for subsequent online matching and triggering.

\begin{itemize} [leftmargin=*, itemindent=0.0em]
\item {\textbf{Time-aware interaction.}} 
For time-related instructions from the user (\eg, ``Remind me to get off in 5 minute''), StreamingProactivity parses the time constraint, trigger condition, and target action and creates a corresponding time-aware reminder node.

\item {\textbf{Event-grounding interaction.}} When the video stream exhibits event cues that satisfy preset conditions (\eg, “a goal is scored”), StreamingProactivity parses the event type and context and creates an event-driven reminder node to record the triggering evidence and response template.
\end{itemize}

\textbf{(2) Proactive response generation.} 
To support proactive perception across multiple domain-specific streaming scenarios, we propose a vision-signal-driven training scheme for proactive perception. This method models state changes in the video stream as a visual-language signal and introduces trigger tokens into the vocabulary to decouple the perception process from downstream tasks. Unlike existing streaming video agents that adopt a unified token design, we customize dedicated trigger tokens for different scenarios to avoid semantic ambiguity, enabling more accurate differentiation of proactive interaction action requirements across scenarios.

\begin{itemize} [leftmargin=*, itemindent=0.0em]
\item {\textbf{Reminder trigger.}} When the visual evidence in the current video chunk matches the trigger conditions defined in a reminder node, such as reaching the specified time point or detecting the target event, proactive interaction is triggered.

\item {\textbf{Response generation.}} StreamingProactivity generates a language response conditioned on the latest visual context and task constraints, such as ``Time is up. Please get ready to get off" or “A goal has been scored. Player No. 1 performed outstandingly."
\end{itemize}

\textbf{(3) Proactive objective evolution.} During real-time interaction, when the user proposes a new proactive objectives online or modifies a previously specified ones, StreamingProactivity either refreshes the goal or evolves the existing one. This online evolution strategy enables an interactive closed loop of monitor trigger $\to$ feedback–$ \to$ evolve $\to$ monitor again.

\begin{figure}
    \centering
    \includegraphics[width=\textwidth]{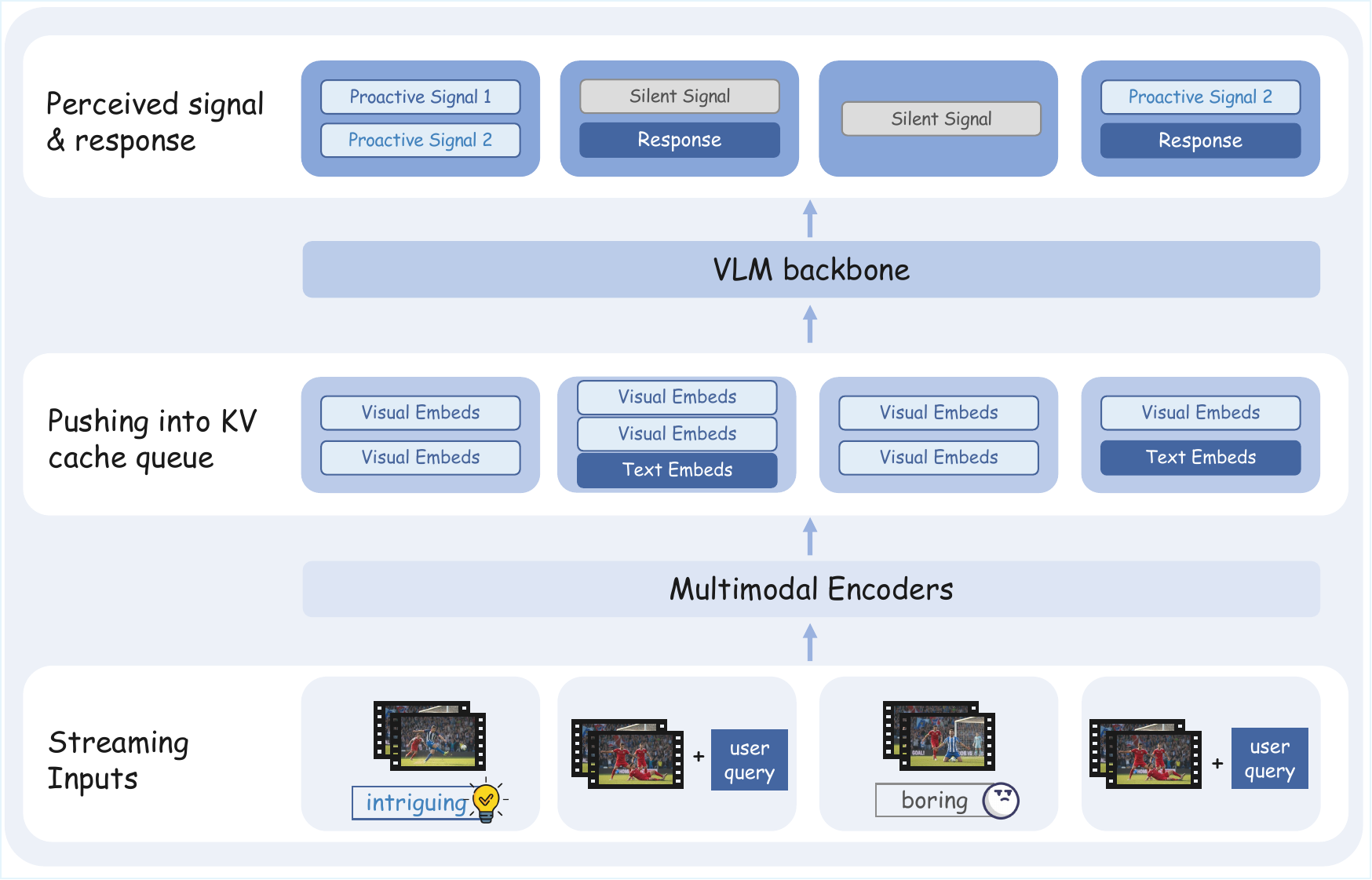}
    \caption{Flowchart of the training-based adaptation pipeline for the StreamingProactivity agent.}
    \label{fig:StreamingProactivity_Training_cropped}
\end{figure}

\subsection{Training Adaption}
To support multiple domain-specific proactive perception scenarios (\eg, continuous in-cabin sensing), we propose a visual-language-driven training scheme for proactive perception agents. This approach models state changes in streaming videos as visual-language signals and introduces trigger tokens into the vocabulary to decouple the perception process from downstream tasks. Unlike existing streaming video agents that adopt a unified trigger-token design \cite{qian2025dispider, xia2025streaming}, we customize scenario-specific trigger tokens to avoid semantic ambiguity, thereby more accurately distinguishing proactive interaction requirements across scenarios. Based on scenario differences, we design two types of proactive inference settings for training adaptation:

\begin{itemize} [leftmargin=*, itemindent=0.0em]
\item \textbf{Silent inference:} The input is purely streaming videos, and the output is one or more proactive signals.
\item \textbf{Non-silent inference:} The input is multimodal streaming data (text, video, and audio), and the output includes both the proactive signals and the agent’s reasoning response.
\end{itemize}

As shown in Fig. \ref{fig:StreamingProactivity_Training_cropped}, StreamingProactivity produces different proactive signals depending on the input type. When there is no user query, the agent performs silent inference: the agent outputs either the \textbf{$\texttt{Silent}$} signal (if no trigger event occurs) or \textbf{$\texttt{Proactive}$} signals (if a trigger event is detected). When a user query is provided, the agent performs non-silent inference: it outputs the corresponding signals and also generates reasoning responses.

Regarding training objectives, under silent inference, the agent only needs to learn a mapping from streaming video to trigger tokens, reformulating the traditional ``detect/classify state changes'' task into a language modeling problem of ``predicting a token sequence.'' Under non-silent inference, beyond learning the video-to-trigger-token mapping, the agent also needs to improve general question-answering capabilities under such scenarios, combining the most recent visual information within the current window with contextual instructions to generate proactive, interactive responses.

In terms of data construction, to enable the agent to learn the mapping from state transition to proactive trigger and response, each streaming video sample needs to be annotated with:

\begin{itemize} [leftmargin=*, itemindent=0.0em]

\item Normal-state segments and changed-state segments (including before/after comparisons) to reduce false triggers.

\item Precise timestamps (event start/end, trigger point, and duration interval) and the corresponding proactive signals.

\item Target response text after triggering, which may include user-visible responses plus structured skill parameters, enabling end-to-end learning from trigger-token prediction to ` "generate response $\to$ invoke skills $\to$ execute actions."

\end{itemize}

The training adaptation approach can achieve the following objectives:
\begin{itemize} [leftmargin=*, itemindent=0.0em]
\item \textbf{Support multiple event types:} The agent can connect to diverse proactive interaction scenarios via a unified visual-signal triggering interface. With scenario-specific signal mappings, it improves the accuracy of multi-event recognition, stabilizes trigger timing, and enhances cross-scenario generalization.

\item \textbf{Customizable events with high recognition accuracy:} By constructing domain-tailored video training data, the training scheme learns the correspondence between events and proactive signals, achieving higher event recognition accuracy than non-training-based approaches.

\item \textbf{Reduced inference overhead:} When multiple proactive interaction requirements coexist (\eg, time-aware requests together with multiple event-aware requests), the model can generate multiple scenario-specific tokens as response signals in a single forward pass, instead of running inference multiple times to produce multiple signals. This reduces inference token overhead while maintaining event recognition accuracy and improves the overall robustness of the framework under concurrent multi-demand settings.
\end{itemize}

With the above design, the StreamingProactivity enables always-on proactive interaction and supports customized proactive interaction scenarios driven by diverse visual state changes.

\section{Scalable Tools and Skills}
\label{sec:tool_and_skill}
To enable multimodal embodied interaction, we equip StreamingClaw with a comprehensive toolbox \cite{shen2024llm,xu2025llm,li2025review,chen2025mindwatcher} and skill library\footnote{https://github.com/anthropics/skills}, thereby completing the final link of the perception–decision–action closed loop. Beyond standard tool compositions, we introduce specialized tools tailored to video understanding and streaming interaction. In addition, while maintaining compatibility with Openclaw skills, we extend an additional set of embodied-interaction skills grounded in real-world application requirements.

\subsection{Basic Toolbox}
\label{sec:toolbox}

To further strengthen the agent’s reasoning capability in streaming video understanding, we introduce a basic toolbox. In addition to conventional agent tools (\eg, \ image magnification and web search tools \cite{chen2025mindwatcher,zheng2025deepeyes,hong2025deepeyesv2,su2025thinking, shen2026evolving}), this toolbox provides standardized video and memory processing primitives and tightly couples them with the agent model via tool-augmented reasoning. From an interface-design perspective, the toolbox adopts a unified data schema for multimodal inputs and tool outputs and exposes a consistent invocation interface supporting both synchronous and asynchronous execution. This design is complemented by a runtime mechanism that seamlessly integrates with a streaming inference stack. Below we detail two tools (\ie, video cut and call memory) as representative examples to illustrate their mechanisms and usage.

\textbf{Video Cut:} Inspired by current works on video understanding \cite{yang2025longvt, jain2025sage,zhang2025thinking}, we design the video cut tool to support fine-grained video perception and semantic understanding. When solving video-centric tasks, the agent proactively invokes this tool whenever it detects that a specific temporal region requires deeper analysis. For example, to answer queries such as “Determine when the girl appears and describe her actions in detail," the agent first predicts a pair of timestamps (start and end times) and passes them to a video cutting module, which extracts a short clip from the original long video for subsequent fine-grained information extraction.
Meanwhile, conditioned on the task context, the agent generates a targeted query instruction specifying the information to be extracted and the analysis focus (\eg, identifying the person, actions, and interaction targets). The cropped clip and the query instruction are then fed into a high-capacity MLLM for in-depth understanding (\eg, Qwen3.5 397B-A17B \cite{qwen3.5} or Qwen3-VL-235B-A22B \cite{qwen3-vl} ). The MLLM’s output is returned as the tool result to the calling agent for downstream decision-making or direct response generation. If the agent determines that the returned information is insufficient, it can iteratively re-invoke the tool with refined temporal boundaries or queries.
Notably, the video cut tool adopts a cascaded pipeline of temporal localization $\to$ fine-grained clipping $\to$ large-model parsing and returns text-only results rather than replaying video content back to the agent. This design substantially reduces context length pressure and achieves an effective balance between computational efficiency and perceptual fidelity.  The specific input and output parameter information of the video cut tool is shown in the following table:
% \textcolor{red}{Fig. \ref{fig:video_cut_cropped} presents a representative example where the video agent correctly instantiates tool parameters and answers the query by reasoning over the returned tool output.

\begin{toolbox}{Tool: Video Cut}
    \textbf{Description:} Given two timestamps, propose the corresponding sub-clips and return the analysis results. \\
    \textbf{Input:} The video and timestamps. \\
    \textbf{Output:} Textual response.  \\
    \textbf{Arguments:}
    \begin{itemize}
        \item \texttt{Query}: Query for video cut tools.
        \item  \texttt{Path}: Path to the video file.
        \item  \texttt{Start time}: Start time in seconds.
        \item  \texttt{End time}: End time in seconds, must be larger than start time.
    \end{itemize}
\end{toolbox}

\textbf{Call Memory:} To better handle long-horizon video question answering, we introduce the call memory tool, which provides StreamingClaw agents with structured access to historical memories. This enables agents to expand their temporal receptive field and perform cross-time comparison, evolution tracking, and causal reasoning.
Faced with a complex input task, the calling agent first engages in strategic decision-making, evaluating whether the current task requires historical memory or can be answered directly based on current information. If the agent decides to invoke the call memory tool, it generates a memory retrieval query. Importantly, this query is not a direct copy of the user prompt. Instead, it is a rewritten retrieval instruction produced after intent decomposition and reasoning. For instance, when the user asks, “What has changed in traffic conditions compared to five minutes ago?” the agent may issue a query to the StreamingMemory, such as “Describe the traffic conditions and key characteristics five minutes ago in detail.” After receiving the retrieved memory (\eg, the road situation five minutes earlier), the video agent compares it against the current working memory (\eg, the real-time road condition), conducts joint reasoning, and produces an accurate differential description for the user. 
The specific input and output parameter information of the call memory tool is shown in the following table:
% \textcolor{red}{As shown in Fig.\ref{fig:call_memory_cropped},} we provide an example of call memory invocation in a streaming setting. 

% \vspace{-30pt}
\begin{toolbox}{Tool: Call Memory}
    \textbf{Description:} Perform memory retrieval for the given query. The input is the user query, and the output is one or more memory nodes, including event summaries, timestamps, and key entities/tags.\\
    \textbf{Input:} Query used to retrieve the memory agent.\\
    \textbf{Output:} Textual response.  \\
    \textbf{Arguments:}
    \begin{itemize}
        \item \texttt{Query}: Query for the memory agent.
    \end{itemize}
\end{toolbox}
% \vspace{-40pt}

Moreover, to improve the reliability, controllability, and generalization of tool usage in complex tasks, we propose a dedicated post-training pipeline for tool-augmented reasoning \cite{jain2025sage,yang2025longvt,open-o3,chen2025mindwatcher,zhang2025thinking}. 
We initialize from state-of-the-art open-source multimodal foundation models (\eg, Qwen3-VL \cite{qwen3-vl}, MindGPT-4ov \cite{mindgpt4ov}) to inherit strong general reasoning and dialogue capabilities and then continue post-training to acquire the key competencies required for tool manipulation.
The pipeline comprises three stages: Supervised Fine-Tuning (SFT), Reinforcement Learning (RL) \cite{deepseekmath,deepseekr1,dpo,jain2025sage, han2025copo}, and Reinforcement Fine-Tuning (RFT) \cite{yang2025longvt,rft}. 
Notably, during the reinforcement learning stage, we introduce a multi-dimensional reward system encompassing outcome rewards, process rewards, and confidence rewards \cite{damani2025beyond,jain2025sage,rft,chen2025mindwatcher}. 
These stages are progressively aligned with our objectives: SFT strengthens video perception/understanding and tool-use capability; RL optimizes tool-calling policies; and RFT stabilizes the learned behaviors and mitigates catastrophic forgetting.
These closed-loop procedures not only consolidate the tool-calling capabilities but also effectively repair the RL-induced knowledge degradation, ultimately yielding a final model that combines strong general reasoning with excellent tool collaboration capabilities.

% \vspace{-40pt}
\subsection{Skill Library}\label{sec:skill}
To improve reusability and extensibility across heterogeneous tasks, StreamingClaw introduces a dedicated skill library for both general-purpose requests and embodied interaction under streaming inputs. The library distills frequently used capabilities into composable, callable skill primitives and enables rapid orchestration and closed-loop execution through a unified interface and scheduling mechanism.

General skills cover common functionalities and remain fully compatible with the OpenClaw ecosystem. They mainly fall into two categories: (1) QA-oriented skills, including open-domain question answering, navigation assistance, weather lookup, and travel recommendations; and (2) entertainment and daily-life skills, such as media playback control, personal memory recall, calendar and reminder management.
Beyond the aforementioned general skills, we further construct a specialized suite of skills tailored specifically for embodied interaction. In the following sections, we provide detailed illustrations using embodied vehicles, embodied robots, and AI wearable devices as representative examples.

\begin{toolbox}{Skill: Embodied Vehicle Driver Monitoring System}
    \textbf{Description:} Monitor the driver in real time via the in-vehicle camera, including dangerous driving behaviors such as fatigued driving (eyes closed, yawning) and distracted driving (head down, using a mobile phone, gaze deviation). \\
    \textbf{Trigger Scenarios:}
    \begin{itemize}
        \item  Driver head down, using mobile phone, line of sight deviated. Trigger fatigue\_state 0.
        \item Driver yawning. Trigger fatigue\_state 1.
        \item  Driver eyes closed. Trigger fatigue\_state 2.
    \end{itemize}
    \textbf{Output Format:} 
    \begin{verbatim}
        {
            "name": "driver_fatigue_warning",
            "required": ["fatigue_state"]
        }
    \end{verbatim}
\end{toolbox}

\textbf{Embodied Vehicle Driver Monitoring System Skill:} This skill leverages in-cabin multimodal sensors to perform real-time driver-state perception and risk assessment. Given continuous streaming video input, the system can reliably detect a range of hazardous behaviors, including fatigued driving (\eg,  prolonged eye closure and yawning) and distracted driving (\eg,  head-down posture, mobile phone usage, and salient gaze deviation). The skill adopts a graded warning policy: eye closure triggers the highest-level fatigue alert (Level 2), yawning triggers a moderate alert (Level 1), and distraction-related behaviors such as head-down posture or phone usage trigger a baseline alert (Level 0). The skill outputs structured data with a standardized fatigue-state level, providing critical decision signals for downstream intervention modules.

\begin{toolbox}{Skill: Embodied Robot Household Care}
    \textbf{Description:} Provide real-time care for family members. When someone is detected falling, initiate active dialogue and take emergency measures. \\
    \textbf{Trigger Scenarios:}
    \begin{itemize}
        \item Detect a person falling down.
        \item  Received a request to make an emergency call.
    \end{itemize}
    \textbf{Output Format:}
    \begin{verbatim}
        {
            "name": "proactive_caring_inquiry",
            "required": ["query"]
        },
        {
            "name": "dial_emergency_number",
            "required": ["phone_num"]
        }
    \end{verbatim}
\end{toolbox}

\textbf{Embodied Robot Household Care Skill:} This skill's goal is to provide real-time safety assurance for family members via multimodal streaming perception. This skill integrates abnormal-behavior detection with emergency response: when a fall is detected or an emergency call request is received, the system automatically triggers a staged intervention pipeline. It first initiates a proactive caring inquiry to assess the user’s condition; if the situation is deemed critical, it escalates to dialing a predefined emergency contact and simultaneously transmits a structured on-scene description. By exposing structured output schemas for both the inquiry instruction and emergency dialing parameters, the skill realizes closed-loop control from passive perception to proactive intervention.

\textbf{AI Glasses Education Tutor Skill:} This skill targets wearable embodied interaction and provides an education tutoring capability built on multimodal streaming perception and real-time dialogue. Under continuous first-person inputs, it supports problem-solving with step-by-step explanations, instant translation, information/literature retrieval, and personalized proactive interaction. Upon receiving queries related to problem solving, translation, or literature search, the system triggers the solve problems tool and emits structured parameters, where the query specifies the request and question type annotates the task category to route execution to an appropriate solver sub-agent. When the user requests sustained follow-ups or proactive engagement, the system invokes create\_proactive\_node, converting the intent into a schedulable proactive node that can later trigger reminders, clarification questions, or task continuation at appropriate times.

\begin{toolbox}{Skill: AI Glasses Education Tutor}
    \textbf{Description:} Provide real-time education tutoring, supporting problem-solving, translation, literature search, and customized proactive interaction. \\
    \textbf{Trigger Scenarios:}
    \begin{itemize}
        \item When receiving inquiries related to problem-solving, translation, and literature search.
        \item  Receive a demand for active interaction.
    \end{itemize}
    \textbf{Output Format:} 
    \begin{verbatim}
        {
            "name": "solve_problems",
            "required": ["query"]
        },
        {
            "name": "create_proactive_node",
            "required": ["query"]
        }
\end{verbatim}   
\end{toolbox}

In terms of runtime design, StreamingClaw follows a skill orchestration procedure similar to OpenClaw. Skills are dynamically loaded on demand: at system initialization, we do not inject the exhaustive specifications and schemas of hundreds of tools/skills into the system prompt. Instead, during inference, the agent selects candidate skills conditioned on the current intent and context and only then loads the corresponding interface definitions and minimal required skill descriptions. This design substantially reduces prompt length and token overhead, mitigates contextual noise, and improves overall runtime efficiency and scalability. 
End-to-end skill execution is driven by StreamingClaw’s agentic loop in a closed-loop manner: the agent first emits a structured skill call with its name and arguments. Subsequently, the runtime scheduler then parses the request and executes the corresponding skill. Next, the execution results, alongside any requisite intermediate evidence, are written back to the dialogue context and working memory. The agent continues reasoning over the updated state, optionally invoking additional skills or retrying with adjusted parameters, until the termination criteria are satisfied and the task is completed.

\section{Conclusion}
StreamingClaw is proposed for scenarios such as embodied intelligence that demand a closed-loop capability of real-time perception–decision–action. To address the limitations of existing video agents in low-latency online inference, long-term memory, and proactive interaction, we build a unified and extensible agent framework. StreamingClaw supports real-time video ingestion and streaming inference for low-latency responses. It also enables cooperative operation by invoking sub-agents for future-event reasoning and proactive interaction decision-making, allowing the system to execute tasks and engage in interactions more proactively under continuous input. To handle information accumulation and utilization in long-duration continuous scenarios, the framework provides a hierarchical multimodal memory that supports memory writing, dynamic evolution, and efficient retrieval. In addition, supported by a tool library and skill library spanning diverse application scenarios, StreamingClaw can reliably translate high-level understanding and decisions into executable actions, improving cross-scenario generality and deployability. Overall, StreamingClaw is compatible with the OpenClaw framework and organically integrates online real-time perception and reasoning, long-term memory, proactive interaction, and extensible action capabilities. It holds promise for applications in embodied intelligence and autonomous driving, advancing the evolution and real-world deployment of general artificial intelligence.

\section{Limitations and Future Work}
StreamingClaw is primarily designed for streaming-video scenarios. It uses MLLMs as agents to support streaming video perception and understanding and provides text-to-speech synthesized outputs. However, at this stage the system still follows a ``vision + text'' input paradigm: speech mainly serves as an output channel, while support remains limited for audio inputs, finer-grained temporal alignment (\eg, audio-visual synchronized understanding), and end-to-end cross-modal joint reasoning and generation.

Future work will evolve toward an omnimodal agent framework: a single unified model that can handle inputs and outputs across video, images, audio, and text, enabling a truly full-duplex, omnimodal closed loop from streaming perception to action. We will also focus on long-horizon temporal modeling, enhanced spatial understanding, and cross-modal alignment mechanisms while optimizing low-latency deployment, advanced memory, and long-horizon tool invocation so that StreamingClaw can achieve stronger embodied interaction capabilities in the real world.

\section*{Contributor List}
All contributors of this work are listed in alphabetical order by their last names.

\paragraph{Core Contributors} Jiawei Chen, Zhe Chen, Chaoqun Du, Maokui He, Wei He, Hengtao Li, Qizhen Li, Zide Liu, Xuhao Pan, Chang Ren,
Xudong Rao, Xintian Shen, Chenfeng Wang, Chengjun Yu, Shengyu Yao, Chunpeng Zhou, Lihao Zheng, Xuhan Zhu, Yufei Zheng

\paragraph{Project Leaders} Hao Ma, Tao Wei, Pengfei Yu

\paragraph{Supervisors} Kun Zhan, Pan Zhou

% Jiawei Chen, Zhe Chen, Chaoqun Du, Maokui He, Wei He, Hengtao Li, Qizhen Li, Zide Liu, Hao Ma, Xuhao Pan, Chang Ren, Xudong Rao, Xintian Shen, Chenfeng Wang, Tao Wei, Chengjun Yu, Pengfei Yu, Shengyu Yao, Chunpeng Zhou, Kun Zhan, Lihao Zheng, Pan Zhou, Xuhan Zhu, Yufei Zheng

\newpage

\bibliographystyle{plainnat}
\bibliography{related_work}

\newpage

\section*{Appendix}

\subsection*{A. Full Example of Tools}
\begin{figure}[h]
    \centering
    \includegraphics[width=\textwidth]{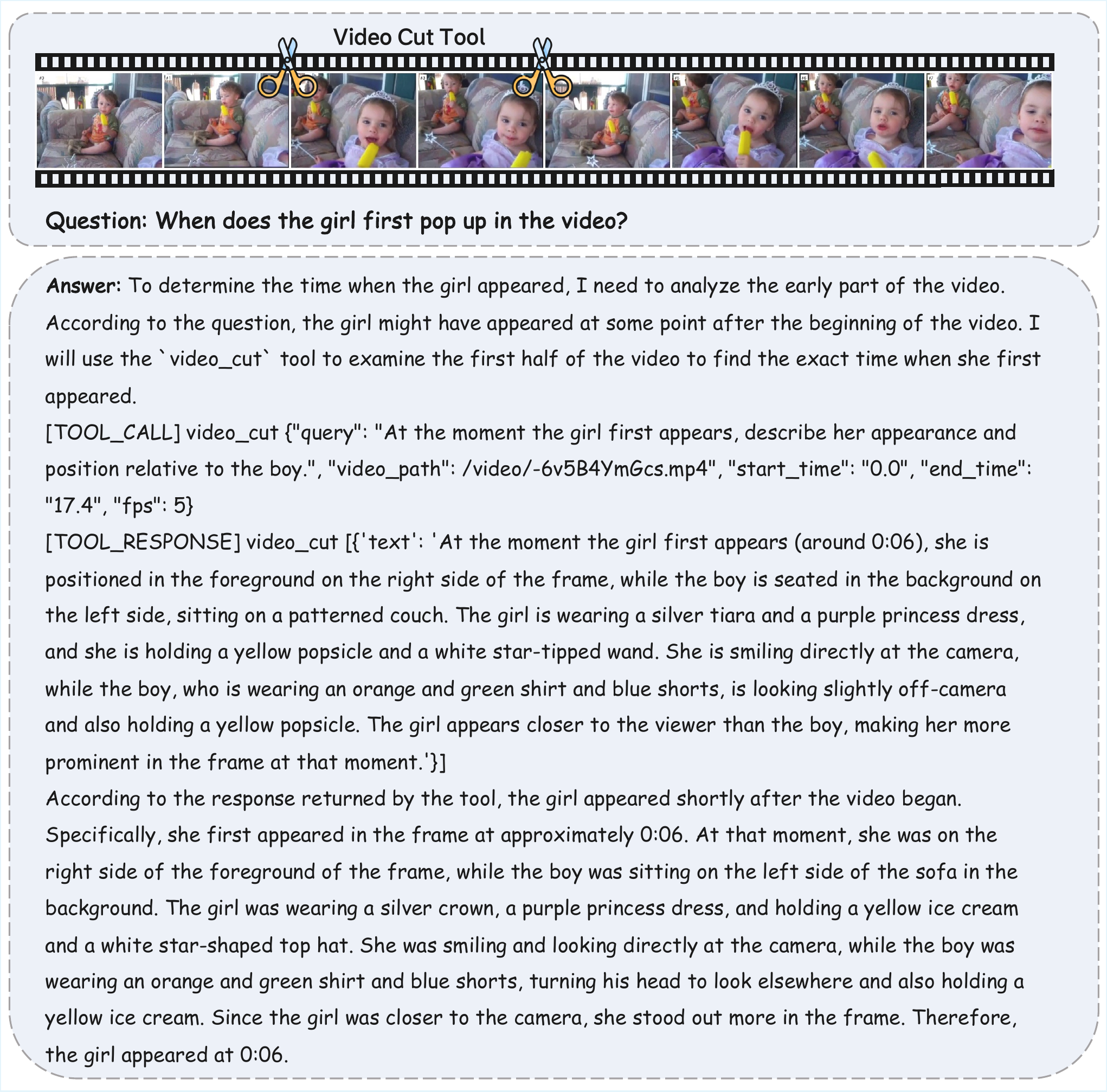}
    \caption{Example of video cut tool invocation.}
    \label{fig:video_cut_cropped}
\end{figure}

\begin{figure}[t]
    \centering
    \includegraphics[width=\textwidth]{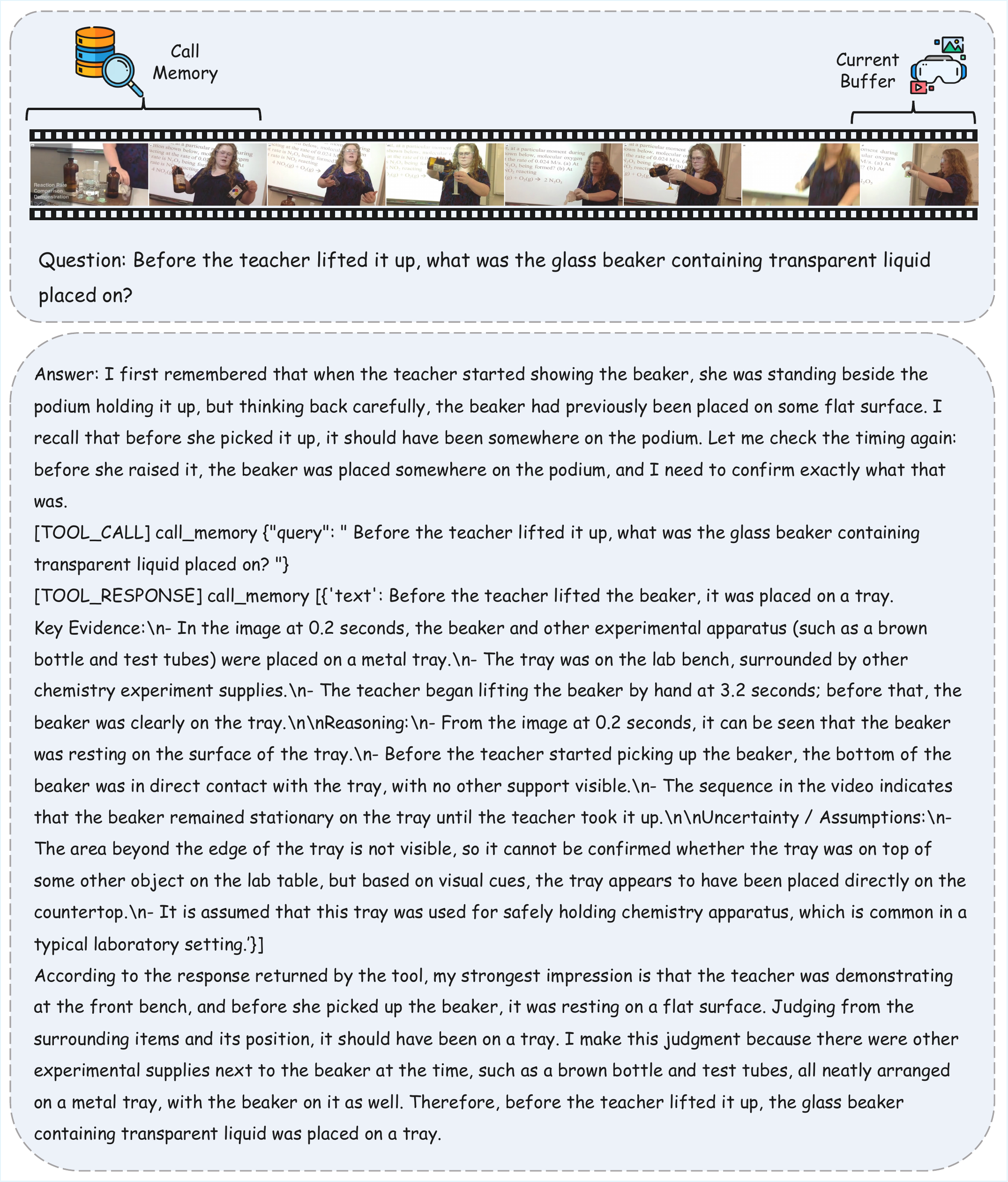}
    \caption{Example of call memory tool invocation.}
    \label{fig:call_memory_cropped}
\end{figure}

In this section, we present complete examples of the tools implemented in StreamingClaw, specifically the video cut tool and the call memory tool. As illustrated in Fig.\ref{fig:video_cut_cropped} and Fig.\ref{fig:call_memory_cropped}. We detail their invocation schemas, runtime execution flow, and how their returned results are incorporated into the agent’s subsequent reasoning.

\newpage

\subsection*{B. Full Example of Skills}
In this section, we present full specifications for three representative embodied interaction skills (discussed in Sec. \ref{sec:skill}) developed within StreamingClaw. 
To demonstrate the broad applicability and robust generalization of StreamingClaw across diverse hardware platforms and complex real-world scenarios, we detail the following: the Embodied Vehicle Driver Monitoring System Skill, designed for real-time safety assessment and intervention; the Embodied Robot Household Care Skill, tailored for proactive domestic assistance and emergency response; and the AI Glasses Education Tutor Skill, engineered to provide context-aware learning support. The following detailed examples illustrate how the agent's high-level cognitive reasoning is seamlessly translated into domain-specific, actionable workflows.
\\
\\

\begin{lstlisting}[style=prompt, caption={Embodied Vehicle Driver Monitoring System Skill}, label={lst:prompt4s_att}]
Embodied Vehicle Driver Monitoring System Skill:
    Description: Monitor the driver in real time via the in-vehicle camera, including dangerous driving behaviors such as fatigued driving (eyes closed, yawning) and distracted driving (head down, using a mobile phone, gaze deviation). 
    
    Trigger Scenarios:
        1. Driver head down, using mobile phone, line of sight deviated. Trigger fatigue\_state 0.
        2. Driver yawning. Trigger fatigue\_state 1.
        3. Driver eyes closed. Trigger fatigue\_state 2.
        
    Output Format:
    {
    "name": "driver_fatigue_warning",
    "parameters": {
      "properties": {
        "fatigue_state": {
          "type": "integer",
          "description": " Driver Fatigue State Level(max 2)",
          "default": 0
        }
      },
      "required": [
        "fatigue_state"
      ]
    }
\end{lstlisting}

\begin{lstlisting}[style=prompt, caption={Embodied Robot Household Care Skill}, label={lst:prompt4s}]
Embodied Robot Household Care Skill:
    Description: Provide real-time care for family members. When someone is detected falling, initiate active dialogue and take emergency measures. 
    Trigger Scenarios:
        1. Detect a person falling down.
        2. Received a request to make an emergency call.
        
    Output Format:
    {
    "name": "proactive_caring_inquiry",
    "parameters": {
      "properties": {
        "query": {
          "type": "string",
          "description": "Provide care and inquiries based on observed situations",
        }
      },
      "required": [
        "query"
      ]
    }
    {
    "name": "dial_emergency_number",
    "parameters": {
      "properties": {
        "phone_num": {
          "type": "string",
          "description": "Emergency contact phone number",
          "default":"123456789"
        },
        "scene_description": {
          "type": "string",
          "description": "Description of the on-site situation",
          "default":"I have detected a person fallen. He requires assistance."
        }
      },
      "required": [
        "phone_num"
      ]
    }
\end{lstlisting}

\begin{lstlisting}[style=prompt, caption={AI Glasses Education Tutor Skill}, label={lst:s}]
AI Glasses Education Tutor Skill:
    Description: Provide real-time education tutoring, supporting problem-solving, translation, literature search, and customized proactive interaction. 
    Trigger Scenarios:
        1. When receiving inquiries related to problem-solving, translation, and literature search.
        2. Receive a demand for active interaction.

    Output Format:
    {
    "name": "solve_problems",
    "parameters": {
      "properties": {
        "query": {
          "type": "string",
          "description": "Problem-solving request",
          "default":"Solve this problem and provide a brief process analysis"
        },
        "question_type": {
          "type": "string",
          "description": "Question type, facilitating the selection of a suitable agent model for solving problems.",
          "default":"STEM"
        }
       },
      "required": [
        "query"
      ]
    }
    {
    "name": "create_proactive_node",
    "parameters": {
      "properties": {
        "query": {
          "type": "string",
          "description": "queries requiring proactive reminders"
        }
      },
      "required": [
        "query"
      ]
    }
\end{lstlisting}

\end{document}